\pdfoutput=1

\documentclass[11pt]{article}

\usepackage{acl}

\usepackage{times}
\usepackage{latexsym}

\usepackage[T1]{fontenc}

\usepackage[utf8]{inputenc}

\usepackage{microtype}

\usepackage{inconsolata}

\usepackage{amsthm,amsmath,amssymb}
 
\usepackage{mathrsfs}
\usepackage{graphicx}
\usepackage{multirow}
\usepackage{svg}
\usepackage{color}
\usepackage{amsmath, bm}

\title{Contextualization Distillation from Large Language Model for Knowledge Graph Completion}

\author{\textbf{Dawei Li\textsuperscript{1}, Zhen Tan\textsuperscript{2}, Tianlong Chen\textsuperscript{3}}, Huan Liu\textsuperscript{2} \\
  \textsuperscript{1}University of California, San Diego \\
  \textsuperscript{2}Arizona State University \\
  \textsuperscript{3}University of North Carolina at Chapel Hill \\
  \texttt{dal034@ucsd.edu,} \texttt{\{ztan36,huanliu\}@asu.edu,} \texttt{tianlong@cs.unc.edu}
}

\begin{document}
\maketitle
\begin{abstract}

While textual information significantly enhances the performance of pre-trained language models (PLMs) in knowledge graph completion (KGC), the static and noisy nature of existing corpora collected from Wikipedia articles or synsets definitions often limits the potential of PLM-based KGC models. To surmount these challenges, we introduce the \textit{Contextualization Distillation} strategy, a versatile plug-in-and-play approach compatible with both discriminative and generative KGC frameworks. Our method begins by instructing large language models (LLMs) to transform compact, structural triplets into context-rich segments. Subsequently, we introduce two tailored auxiliary tasks\textemdash reconstruction and contextualization\textemdash allowing smaller KGC models to assimilate insights from these enriched triplets. Comprehensive evaluations across diverse datasets and KGC techniques highlight the efficacy and adaptability of our approach, revealing consistent performance enhancements irrespective of underlying pipelines or architectures. Moreover, our analysis makes our method more explainable and provides insight into generating path selection, as well as the choosing of suitable distillation tasks. All the code and data in this work will be released at https://github.com/David-Li0406/Contextulization-Distillation

\end{abstract}

\section{Introduction}
\label{Introduction}
Knowledge graph completion (KGC) is a fundamental task in natural language processing (NLP), aiming at unveiling hidden insights within diverse knowledge graphs to explore novel knowledge patterns.
Traditional KGC methods~\cite{nickel2011three,bordes2013translating} typically predict the missing part of the triplets by learning the representation of each entity and relation based on their structural information.
However, such embedding-based methods tend to overlook the rich textual information of the knowledge graph.
Therefore, pre-trained language models (PLMs) have been introduced to KGC and achieved promising results~\cite{kenton2019bert,xie2022discrimination}.

\begin{figure}[!h]
    \centering
    \includegraphics[width=7.7cm]{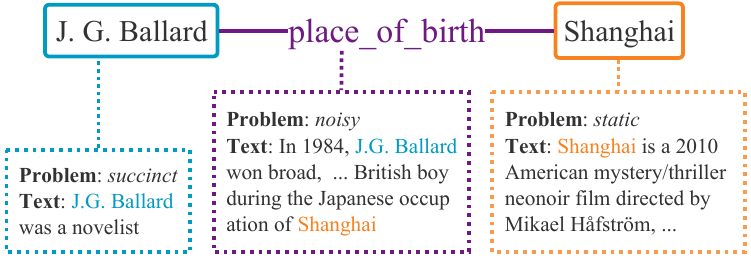}
    \caption{An example to illustrate the limitations of the current textual information for KGC.}
    \label{bad_case}
\end{figure}

\begin{table}[t]\small
\centering
\setlength{\tabcolsep}{2.0mm}{
\begin{tabular}{l|ccc}
\hline
\multicolumn{1}{l|}
{Methods} & H@1           & H@3           & H@8/10        \\ \hline
ChatGPT-1-shot                & 15.6          & 17.6          & 19.6          \\ 
PaLM2-1-shot                  & 15.7          & 20.8          & 25.4          \\ 
KG-S2S~\cite{chen2022knowledge}                 & \textbf{28.5} & \textbf{38.8} & \textbf{49.3} \\ \hline
\end{tabular}}
\caption{ChatGPT and PaLM2's unsatisfactory performance on the test set of FB15k-237N compared to a smaller KGC model, KG-S2S~\cite{chen2022knowledge}.}
\label{preliminary}
\end{table}

While it has been well-discovered that textual information can be beneficial for PLM-based KGC models~\cite{yao2019kg,wang2021kepler,chen2022knowledge,li2022c3kg,chen2023dipping}, prior attempts to augment KGC models with textual data from Wikipedia article~\cite{zhong2015aligning} or synsets definitions~\cite{yao2019kg} have encountered certain limitations:
(\textit{i}) Entity descriptions, often succinct and static, may inhibit the formation of a comprehensive understanding of entities within KGC models. (\textit{ii}) The incorporation of triplet descriptions, albeit potentially enriching, can introduce substantial noise, particularly when derived through automatic entity alignment~\cite{sun2020colake}.
Figure~\ref{bad_case} demonstrates an example to illustrate the aforementioned limitations.
The description for the head ``\emph{J. G. Ballard}'' is limited and for the tail ``\emph{Shanghai}'', it mistakenly uses the definition of the movie also named ``Shanghai''.
Also, while the two entities show up in the triplet description, it falls short in conveying the semantic essence of the relation ``\emph{place\_of\_birth}''.

In light of these limitations, our attention shifts to Large Language Models (LLMs)~\cite{brown2020language,zhang2022opt,anil2023palm,touvron2023llama}, renowned for their capability in generating articulate and high-quality data~\cite{dai2023chataug,shridhar2023distilling,zheng2023augesc}. Our exploration commences with a scrupulous evaluation of LLMs, such as ChatGPT and PaLM2, in KGC, benchmarking them across several esteemed KGC datasets~\cite{dettmers2018convolutional,garcia2018learning,mahdisoltani2013yago3}. 
Utilizing 1-shot In-Context Learning (ICL), we deduce missing heads or tails in triplets and report evaluation metrics.
It reveals a significant performance discrepancy of two LLMs in comparison to KG-S2S~\cite{chen2022knowledge} despite its reliance on a smaller foundational model, T5-base~\cite{raffel2020exploring}. This insight propels us toward the conclusion that direct utilization of LLMs for KGC tasks, while intuitive, is outperformed by the fine-tuning of more diminutive, specialized KGC models. This observation aligns with findings from~\citet{liang2022holistic,sun2023head,zhao2023survey,wei2023kicgpt}, which highlighted the limitations of LLMs in knowledge-centric tasks.
Experiment results and analysis on more KGC datasets can be found in Appendix~\ref{Large Language Model Perform on KGC}.


To optimally harness LLMs for KGC, we draw inspiration from recent works~\cite{xiang2022asdot,kim2022soda} and introduce a novel approach, \textit{Contextualization Distillation}. 
Contextualization Distillation first extracts descriptive contexts from LLMs with well-designed prompts, thereby securing dynamic, high-quality context for each entity and triplet.
Subsequent to this, two auxiliary tasks are proposed to train smaller KGC models with these informative, descriptive contexts.
The plug-in-and-play characteristic of our contextualization distillation enables us to apply and evaluate it on various KGC datasets and baseline models. Through extensive experiments, we affirm that Contextualization Distillation consistently enhances the performance of smaller KGC models, irrespective of architectural and pipeline disparities. Additionally, we provide an exhaustive analysis of each step of Contextualization Distillation, encouraging further insights and elucidations.


The contributions of this work can be summarized into three main aspects:
\begin{itemize}
    \item We identify the constraints of the current corpus for PLMs-based KGC models and introduce a plug-in-and-play approach, Contextualization Distillation, to enhance smaller KGC models with extracted rationale from LLMs.
    \item We conduct extensive experiments across several widely recognized KGC datasets and utilize various baseline models. Through these experiments, we validate the effectiveness of Contextualization Distillation in consistently improving smaller KGC models.
    \item We delve into a comprehensive analysis of our proposed method and provide valuable insights and guidance on generating path selection for distillation, as well as the selection of suitable distillation tasks.
\end{itemize}

\section{Related Work}
\subsection{Knowledge Graph Completion}

Traditional KGC methods~\cite{nickel2011three,bordes2013translating} involve embedding entities and relations into a representation space.
In pursuit of a more accurate depiction of entity-relation pairs, different representation spaces~\cite{trouillon2016complex,xiao2016transg} have been proposed considering various factors, e.g., differentiability and calculation possibility~\cite{ji2021survey}.
During training, two primary objectives emerge to assign higher scores to true triplets than negative ones: 1) Translational distance methods gauge the plausibility of a fact by measuring the distance between the two entities under certain relations~\cite{lin2015learning,wang2014knowledge}; 2) Semantic matching methods compute the latent semantics of entities and relations~\cite{yang2015embedding,dettmers2018convolutional}.

\begin{figure*}[!t]

    \centering
    \includegraphics[width=16cm]{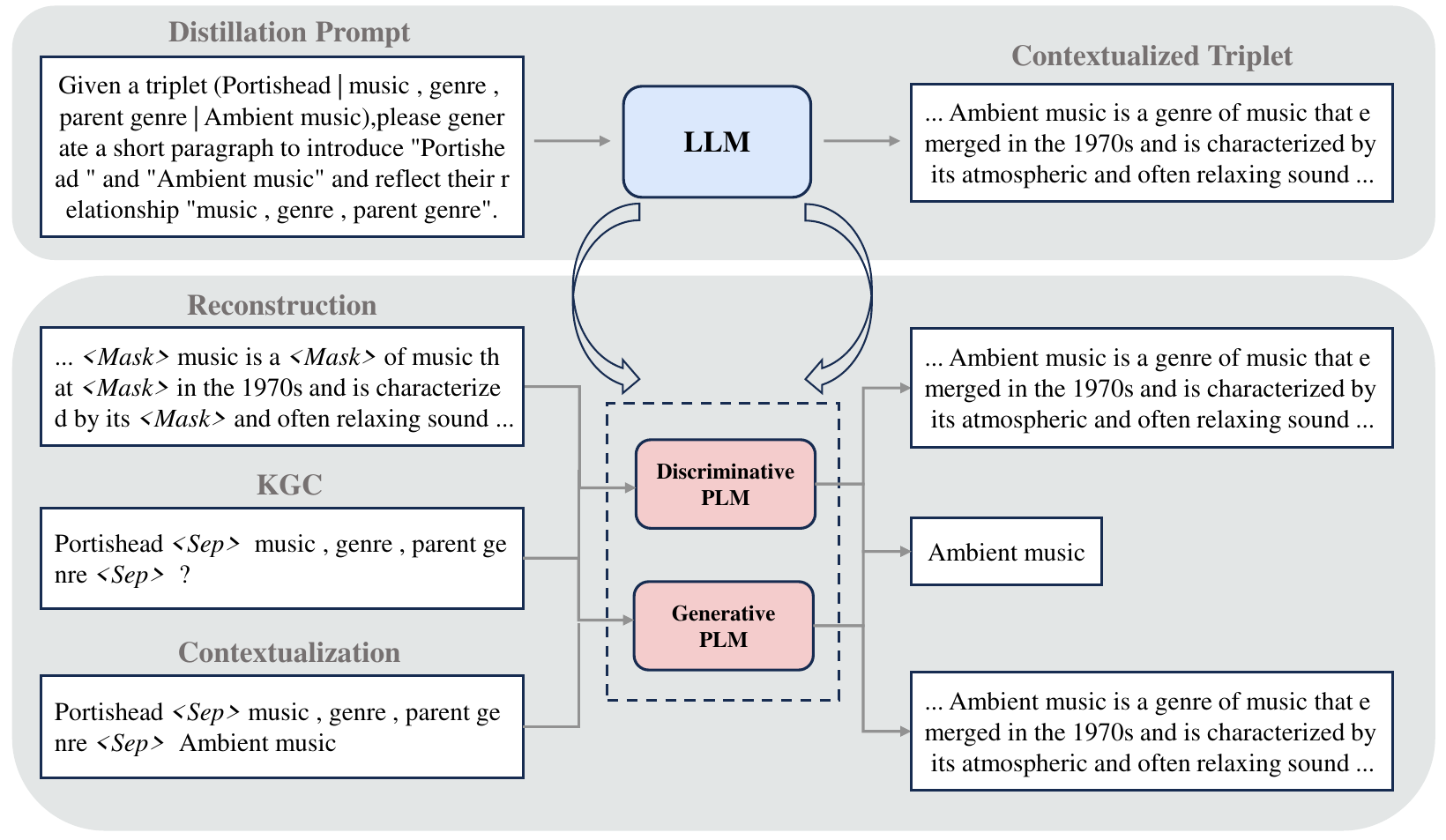}
    \caption{An overview pipeline of our Contextualization Distillation. We first extract descriptive contexts from LLMs (Section~\ref{Extract Descriptive Context from LLMs}). Then, two auxiliary tasks, reconstruction (Section~\ref{Reconstruction}) and contextualization (Section~\ref{Contextualization}) are designed to train the smaller KGC models with the contextualized information.}
    \label{overview pipeline}
\end{figure*}

To better utilize the rich textual information of knowledge graphs, PLMs have been introduced in KGC.
\citet{yao2019kg} first propose to use BERT~\cite{kenton2019bert} to encode the entity and relation's name and adopt a binary classifier to predict the validity of given triplets.
Following them,~\citet{wang2021structure} leverage the Siamese network to encode the head-relation pair and tail in a triplet separately, aiming to reduce the time cost and make the inference scalable.
\citet{lv2022pre} convert each triple and its textual information into natural prompt sentences to fully inspire PLMs' potential in the KGC task.
\citet{chen2023dipping} design a conditional soft prompts framework to maintain a balance between structural information and textual knowledge in KGC.
Recently, there are also some works trying to leverage generative PLMs to perform KGC in a sequence-to-sequence manner and achieve promising results~\cite{xie2022discrimination,saxena2022sequence,chen2022knowledge}. 

\subsection{Distillation from LLMs}

Knowledge distillation has proven to be an effective approach for transferring expertise from larger, highly competent teacher models to smaller, affordable student models~\cite{buciluǎ2006model,hinton2015distilling,beyer2022knowledge}.
With the emergence of LLMs, a substantial body of research has concentrated on distilling valuable insights from these LLMs to enhance the capabilities of smaller PLMs.
One of the most common methods is to prompt LLMs to explain their predictions and then use such rationales to distill their reasoning abilities into smaller models~\cite{wang2022pinto,ho2022large,magister2022teaching,hsieh2023distilling,shridhar2023distilling}.
Distilling conversations from LLMs is another cost-effective method to build new dialogue datasets~\cite{kim2022prosocialdialog,chen2023places,kim2022soda} or augment existing ones~\cite{chen2022weakly,zhou2022reflect,zheng2023augesc}.
There are also some attempts~\cite{marjieh2023language,zhang2023huatuogpt} that focus on distilling domain-specific knowledge from LLMs for various downstream applications.

Several recent studies have validated the contextualization capability of LLMs to convert structural data into raw text.
Among them,~\citet{xiang2022asdot} convert triplets in the data-to-text generation dataset into their corresponding descriptions to facilitate disambiguation.
\citet{kim2022soda} design a pipeline for synthesizing a dialogue dataset by distilling conversations from LLMs, enhanced with a social commonsense knowledge graph.
By contrast, we are the first to leverage descriptive context generated by LLMs as an informative auxiliary corpus to the KGC models.

\section
{Contextualization Distillation}

In this section, we first illustrate how we curate prompts to extract the descriptive context of each triplet from the LLM.
Subsequently, we design a multi-task framework, together with two auxiliary tasks\textemdash reconstruction and contextualization\textemdash to train smaller KGC models with these high-quality context corpus.
The overview pipeline of our method is illustrated in Figure~\ref{overview pipeline}.

\subsection{Extract Descriptive Context from LLMs}
\label{Extract Descriptive Context from LLMs}

\begin{figure}[!t]
    \centering
    \includegraphics[width=8cm]{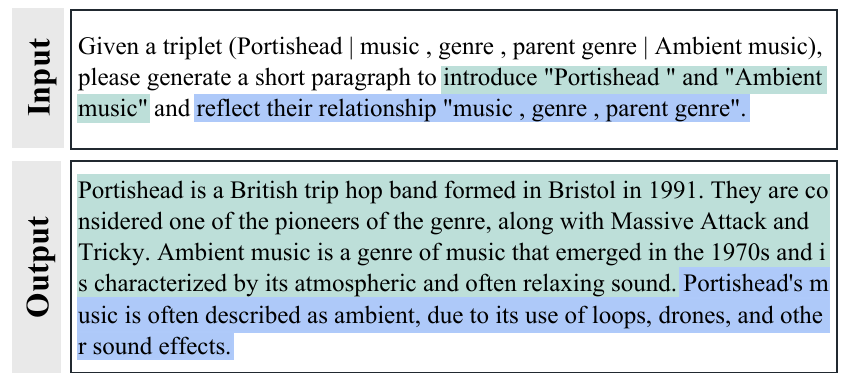}
    \caption{An example contains our instruction to LLMs and the generated descriptive context. We use green to highlight entity description prompt/ generation result and blue to highlight triplet description prompt/ generation result.}
    \label{prompt}
\end{figure}

Recent studies have highlighted the remarkable ability of LLMs to contextualize structural data and transform it into context-rich segments~\cite{xiang2022asdot,kim2022soda}.
Here we borrow their insights and extract descriptive context from LLMs to address the limitations of the existing KGC corpus we mentioned in Section~\ref{Introduction}.

In particular, we focus on two commonly employed types of descriptions prevalent in prior methodologies: entity description (ED)~\cite{yao2019kg,chen2022knowledge} and triplet description (TD)~\cite{sun2020colake}.
Entity description refers to the definition and description of individual entities, while triplet description refers to a textual segment that reflects the specific relationship between two entities within a triplet.
Given triplets of a knowledge graph $t_i \in T$, we first curate prompt $p_i$ for the $i^{th}$ triplet by filling the pre-defined template:
\begin{equation}
    p_i = {\rm Template}(h_i, r_i, t_i),
\end{equation}
where $h_i$, $r_i$, $t_i$ are the head entity, relation, and tail entity of the $i^{th}$ triplet.
Then, we use $p_i$ as the input to prompt the LLM to generate the descriptive context $c_i$ for each triplet:
\begin{equation}
   c_i = {\rm LLM}(p_i),
\end{equation}

\subsection{Generating Path}

Without loss of generalization, we consider different generating paths to instruct the LLMs to generate textual information and conduct an ablation study in Section~\ref{Ablation Study on Generating Path}.
All the generating paths we adopt are as follows:


$\bm{T \longrightarrow (ED,TD)}$ generates both entity description and triplet description at one time. As Figure~\ref{prompt} shows, this is the context generating path we use in the main experiment.

$\bm{T \longrightarrow ED}$ curates prompt to instruct the LLM to generate the entity description only.

$\bm{T \longrightarrow TD}$ curates prompt to instruct the LLM to generate the triplet description only.

$\bm{T \longrightarrow RA}$ prompts the LLM to generate rationale rather than descriptive context.

$\bm{T \longrightarrow ED \longrightarrow TD}$ produces entity description and triplet description in a two-step way. The final descriptive context is obtained by concatenating the two segments of text.

We also give further details and examples of our prompt in Appendix~\ref{Additional Case Study}.


\subsection{Multi-task Learning with Descriptive Context}

Different PLM-based KGC models adopt diverse loss functions and pipeline architectures~\cite{yao2019kg,chen2022knowledge, xie2022discrimination,chen2023dipping}.
\textbf{To ensure the compatibility of our Contextualization Distillation to be applied in various PLM-based KGC methods}, we design a multi-task learning framework for these models to learn from both the KGC task and auxiliary descriptive context-based tasks.
For the auxiliary tasks, we design \emph{reconstruction} (Section~\ref{Reconstruction}) and \emph{contextualizatioin} (Section~\ref{Contextualization}) for discriminative and generative KGC models respectively.

\subsubsection{Reconstruction}
\label{Reconstruction}

The reconstruction task aims to train the model to restore the corrupted descriptive contexts.
For the discriminative KGC models, we follow the implementation of~\citet{kenton2019bert} and use masked language modeling (MLM).
Previous studies have validated that \textbf{such auxiliary self-supervised tasks in the domain-specific corpus can benefit downstream applications}~\cite{han2021fine,wang2021kepler}.

To be specific, MLM randomly identifies 15\% of the tokens within the descriptive context. Among these tokens, 80\% are tactically concealed with the special token ``$<Mask>$'', 10\% are seamlessly substituted with random tokens, while the remaining 10\% keep unchanged.
For each selected token, the objective of MLM is to restore the original content at that particular position, achieved through the cross-entropy loss.
The aforementioned process can be formally expressed as follows:
\begin{equation}
    c_i^{'} = {\rm MLM}(c_i),
\end{equation}
\begin{equation}
    \mathcal{L}_{rec} = \frac{1}{N} \sum_{i=1}^{N} \ell (f(c_i^{'}),c_i)
\end{equation}

The final loss of discriminative KGC models is the combination of the KGC loss\footnote{We give the illustration of the discriminative KGC models we used in Appendix~\ref{Discriminative KGC Pipelines}} and the proposed reconstruction loss:
\begin{equation}
    \mathcal{L}_{dis} = \mathcal{L}_{kgc} + \alpha \cdot \mathcal{L}_{rec},
\end{equation}
where $\alpha$ is a hyper-parameter to control the ratios between the two losses.

\subsubsection{Contextualization}
\label{Contextualization}

The objective of contextualization is to instruct the model in generating the descriptive context $c_i$ when provided with the original triplet $t_i={h,r,t}$. Compared with reconstruction, \textbf{contextualization demands a more nuanced and intricate ability from PLM}.
It necessitates the PLM to precisely grasp the meaning of both entities involved and the inherent relationship that binds them together, to generate fluent and accurate descriptions.

Specifically, we concatenate head, relation and tail with a special token ``$<Sep>$'' as input:
\begin{equation}
    I_i = {\rm Con}(h_i,<Sep>,r_i,<Sep>,t_i)
\end{equation}
Then, we input them into the generative PLM and train the model to generate descriptive context $c_i$ using the cross-entropy loss:
\begin{equation}
    \mathcal{L}_{con} = \frac{1}{N} \sum_{i=1}^{N} \ell (f(I_i),c_i)
\end{equation}

The final loss of generative KGC models is the combination of the KGC loss\footnote{We give the illustration of the generative KGC models we used in Appendix~\ref{Generative KGC Pipelines}} and the proposed contextualization loss:
\begin{equation}
    \mathcal{L}_{gen} = \mathcal{L}_{kgc} + \alpha \cdot \mathcal{L}_{con}
\end{equation}

For generative KGC models, it is also applicable to apply reconstruction as the auxiliary task.
We have done an ablation study in Section~\ref{Ablation Study on Generative KGC Models} to examine the effectiveness of each auxiliary task on generative KGC models.

\section{Experiment}
\begin{table*}[t!]\small
\centering
\begin{tabular}{lcccccccc}
\hline
\multirow{2}{*}{}   & \multicolumn{4}{c}{WN18RR}                                                              & \multicolumn{4}{c}{FB15k-237N}                                                          \\ \cline{2-9} 
                    & \multicolumn{1}{c}{MRR}  & \multicolumn{1}{c}{H@1}  & \multicolumn{1}{c}{H@3}  & H@10 & \multicolumn{1}{c}{MRR}  & \multicolumn{1}{c}{H@1}  & \multicolumn{1}{c}{H@3}  & H@10 \\ \hline
\small{\emph{\textbf{Traditional Methods}}} & \multicolumn{1}{c}{}     & \multicolumn{1}{c}{}     & \multicolumn{1}{c}{}     &      & \multicolumn{1}{c}{}     & \multicolumn{1}{c}{}     & \multicolumn{1}{c}{}     &      \\ 
TransE*~\cite{bordes2013translating}              & \multicolumn{1}{c}{24.3} & \multicolumn{1}{c}{4.3}  & \multicolumn{1}{c}{44.1} & 53.2 & \multicolumn{1}{c}{25.5} & \multicolumn{1}{c}{15.2} & \multicolumn{1}{c}{30.1} & 45.9 \\ 
DisMult*~\cite{yang2015embedding}             & \multicolumn{1}{c}{44.4} & \multicolumn{1}{c}{41.2} & \multicolumn{1}{c}{47.0} & 50.4 & \multicolumn{1}{c}{20.9} & \multicolumn{1}{c}{14.3} & \multicolumn{1}{c}{23.4} & 33.0 \\ 
ComplEx*~\cite{trouillon2016complex}             & \multicolumn{1}{c}{44.9} & \multicolumn{1}{c}{40.9} & \multicolumn{1}{c}{46.9} & 53.0 & \multicolumn{1}{c}{24.9} & \multicolumn{1}{c}{18.0} & \multicolumn{1}{c}{27.6} & 38.0 \\ 
ConvE*~\cite{dettmers2018convolutional}               & \multicolumn{1}{c}{45.6} & \multicolumn{1}{c}{41.9} & \multicolumn{1}{c}{47.0} & 53.1 & \multicolumn{1}{c}{27.3} & \multicolumn{1}{c}{19.2} & \multicolumn{1}{c}{30.5} & 42.9 \\ 
RotatE*~\cite{sun2018rotate}              & \multicolumn{1}{c}{47.6} & \multicolumn{1}{c}{42.8} & \multicolumn{1}{c}{49.2} & 57.1 & \multicolumn{1}{c}{27.9} & \multicolumn{1}{c}{17.7} & \multicolumn{1}{c}{32.0} & 48.1 \\ 
CompGCN*~\cite{vashishth2019composition}             & \multicolumn{1}{c}{47.9} & \multicolumn{1}{c}{44.3} & \multicolumn{1}{c}{49.4} & 54.6 & \multicolumn{1}{c}{31.6} & \multicolumn{1}{c}{23.1} & \multicolumn{1}{c}{34.9} & 48.0 \\ 
\hline
\small{\emph{\textbf{PLMs-based Methods}}}   & \multicolumn{1}{c}{}     & \multicolumn{1}{c}{}     & \multicolumn{1}{c}{}     &      & \multicolumn{1}{c}{}     & \multicolumn{1}{c}{}     & \multicolumn{1}{c}{}     &      \\ 
MTL-KGC*~\cite{kim2020multi}             & \multicolumn{1}{c}{33.1} & \multicolumn{1}{c}{20.3} & \multicolumn{1}{c}{38.3} & 59.7 & \multicolumn{1}{c}{24.1} & \multicolumn{1}{c}{16.0} & \multicolumn{1}{c}{28.4} & 43.0 \\ 
StAR*~\cite{wang2021structure}                & \multicolumn{1}{c}{40.1} & \multicolumn{1}{c}{24.3} & \multicolumn{1}{c}{49.1} & \bf{70.9} & \multicolumn{1}{c}{-}    & \multicolumn{1}{c}{-}    & \multicolumn{1}{c}{-}    & -    \\ 
PKGC*~\cite{lv2022pre}                & \multicolumn{1}{c}{-}    & \multicolumn{1}{c}{-}    & \multicolumn{1}{c}{-}    & -    & \multicolumn{1}{c}{30.7} & \multicolumn{1}{c}{23.2} & \multicolumn{1}{c}{32.8} & 47.1 \\ 
KGT5*~\cite{saxena2022sequence}                & \multicolumn{1}{c}{50.8} & \multicolumn{1}{c}{48.7} & \multicolumn{1}{c}{-}    & 54.4 & \multicolumn{1}{c}{-}    & \multicolumn{1}{c}{-}    & \multicolumn{1}{c}{-}    & -    \\ 
\hline
\small{\emph{\textbf{Our Implementation}}}   & \multicolumn{1}{c}{}     & \multicolumn{1}{c}{}     & \multicolumn{1}{c}{}     &      & \multicolumn{1}{c}{}     & \multicolumn{1}{c}{}     & \multicolumn{1}{c}{}     &      \\ 
KG-BERT~\cite{yao2019kg}             & \multicolumn{1}{c}{21.6} & \multicolumn{1}{c}{4.1}  & \multicolumn{1}{c}{30.2} & 52.4 & \multicolumn{1}{c}{20.3} & \multicolumn{1}{c}{13.9} & \multicolumn{1}{c}{20.1} & 40.3 \\ 
KG-BERT-CD      & \multicolumn{1}{c}{30.3} & \multicolumn{1}{c}{16.5} & \multicolumn{1}{c}{35.4} & 60.2 & \multicolumn{1}{c}{25.0} & \multicolumn{1}{c}{17.2} & \multicolumn{1}{c}{26.6} & 45.5 \\ 
GenKGC~\cite{xie2022discrimination}              & \multicolumn{1}{c}{-}    & \multicolumn{1}{c}{28.6} & \multicolumn{1}{c}{44.4} & 52.4 & \multicolumn{1}{c}{-}    & \multicolumn{1}{c}{18.7} & \multicolumn{1}{c}{27.3} & 33.7 \\ 
GenKGC-CD       & \multicolumn{1}{c}{-}    & \multicolumn{1}{c}{29.3} & \multicolumn{1}{c}{45.6} & 53.3 & \multicolumn{1}{c}{-}    & \multicolumn{1}{c}{20.4} & \multicolumn{1}{c}{29.3} & 34.9 \\ 
KG-S2S~\cite{chen2022knowledge}              & \multicolumn{1}{c}{57.0} & \multicolumn{1}{c}{52.5} & \multicolumn{1}{c}{59.7} & 65.4 & \multicolumn{1}{c}{35.4} & \multicolumn{1}{c}{28.5} & \multicolumn{1}{c}{38.8} & 49.3 \\ 
KG-S2S-CD       & \multicolumn{1}{c}{\bf{57.6}} & \multicolumn{1}{c}{\bf{52.6}} & \multicolumn{1}{c}{\bf{60.7}} & 67.2 & \multicolumn{1}{c}{35.9} & \multicolumn{1}{c}{\bf{28.9}} & \multicolumn{1}{c}{39.4} & 50.2 \\ 
CSProm-KG~\cite{chen2023dipping}           & \multicolumn{1}{c}{55.2} & \multicolumn{1}{c}{50.0} & \multicolumn{1}{c}{57.2} & 65.7 & \multicolumn{1}{c}{36.0} & \multicolumn{1}{c}{28.1} & \multicolumn{1}{c}{39.5} & 51.1 \\ 
CSProm-KG-CD    & \multicolumn{1}{c}{55.9} & \multicolumn{1}{c}{50.8} & \multicolumn{1}{c}{57.8} & 66.0 & \multicolumn{1}{c}{\bf{37.2}} & \multicolumn{1}{c}{28.8} & \multicolumn{1}{c}{\bf{41.0}} & \bf{53.0} \\ \hline
\end{tabular}
\caption{Experiment results on WN18RR and FB15k-237. * denotes results we take from~\citet{chen2022knowledge}. Methods suffixed with "-CD" indicate the baseline models with our Contextualization Distillation applied. The best results of each metric are in bold.}
\label{Main Result}
\end{table*}

In this section, we apply our Contextualization Distillation across a range of PLM-based KGC baselines.
We compare our enhanced model with our approach against the vanilla models using several KGC datasets.
Additionally, we do further analysis of each component in our contextualized distillation and make our method more explainable by conducting case studies.

\subsection{Experimental Settings}
\label{Experimental Settings}

\paragraph{Datasets}
We use WN18RR~\cite{dettmers2018convolutional} and FB15k-237N~\cite{lv2022pre} in our experiment.
WN18RR serves as an enhanced version of its respective counterparts, WN18~\cite{bordes2013translating}. The improvements involve the removal of all inverse relations to prevent potential data leakage.
For FB15K-237N, it's a refine version of FB15k~\cite{bordes2013translating}, by eliminating concatenated relations stemming from Freebase mediator nodes~\cite{akrami2020realistic} to avoid Cartesian production relation issues.

\paragraph{Baselines}
we adopt several PLM-based KGC models as baselines and apply the proposed Contextualization Distillation to them.
\textbf{KG-BERT}~\cite{yao2019kg} is the first to suggest utilizing PLMs for the KGC task.
we also consider \textbf{CSProm-KG}~\cite{chen2023dipping}, which combines PLMs with traditional Knowledge Graph Embedding (KGE) models, achieving a balance between efficiency and performance in KGC.
In addition to these discriminative models, we also harness generative KGC models.
\textbf{GenKGC}~\cite{xie2022discrimination} is the first to accomplish KGC in a sequence-to-sequence manner, with a fine-tuned BART~\cite{lewis2020bart} as its backbone.
Following them, \textbf{KG-S2S}~\cite{chen2022knowledge} adopt soft prompt tuning and lead to a new SOTA performance among the generative KGC models.

\paragraph{Implementation details}
All our experiments are conducted on a single GPU (RTX A6000), with CUDA version 11.1.
We use PaLM2-540B\cite{anil2023palm} as the large language model to distill descriptive context.
We tune the Contextualization Distillation hyper-parameter $\alpha \in \{0.1, 0.5, 1.0\}$.
We follow the hyper-parameter settings in the original papers to reproduce each baseline's result.
For all datasets, we follow the previous works~\cite{chen2022knowledge,chen2023dipping} and report Mean Reciprocal Rank (MRR), Hits@1, Hits@3 and Hits@10.
More details about our experiment implementation and dataset statistics are shown in Appendix~\ref{Additional Implementation Details}.

\subsection{Main Result}

Table~\ref{Main Result} displays the results of our experiments on WN18RR and FB15k-237N.
We observe that our Contextualization Distillation consistently enhances the performance of all baseline methods, regardless of whether they are based on generative or discriminative models.
This unwavering improvement demonstrates \textbf{the robust generalization and compatibility of our approach across various PLMs-based KGC methods}.

Additionally, some baselines we choose to implement our Contextualization Distillation also utilize context information.
For example, both KG-BERT and CSProm-KG adopt entity descriptions to enhance entity embedding representation.
Nevertheless, our approach manages to deliver additional improvements to these context-based baselines.
Among them, it is worth noting that the application of our approach to KG-BERT achieves an overall 31.7\% enhancement in MRR.
All these findings lead us to the conclusion that \textbf{Contextualization Distillation is not only compatible with context-based KGC models but also capable of further enhancing their performance}.

\subsection{Ablation Study on Generating Path}
\label{Ablation Study on Generating Path}

\begin{table}[h]
\centering
\begin{tabular}{lccc}
\hline
\multirow{2}{*}{Paths}                                               & \multicolumn{3}{c}{FN15k-237N}                                                                                              \\ \cline{2-4} 
                                                                                 & \multicolumn{1}{c}{H@1}           & \multicolumn{1}{c}{H@3}           & H@10          \\ \hline
-                                                                            & \multicolumn{1}{c}{18.7}          & \multicolumn{1}{c}{27.3}          & 33.7          \\
\begin{tabular}[c]{@{}l@{}}  $T \longrightarrow ED$ \end{tabular}       &  \multicolumn{1}{c}{20.0} & \multicolumn{1}{c}{28.9} &    34.5       \\
\begin{tabular}[c]{@{}l@{}}  $T \longrightarrow TD$ \end{tabular}       & \multicolumn{1}{c}{20.1} & \multicolumn{1}{c}{29.0} &    34.6       \\
\begin{tabular}[c]{@{}l@{}}  $T \longrightarrow RA$ \end{tabular}       & \multicolumn{1}{c}{19.4} & \multicolumn{1}{c}{28.2} & 34.2          \\ 
\begin{tabular}[c]{@{}l@{}} $T \longrightarrow ED \longrightarrow TD$ \end{tabular}       & \multicolumn{1}{c}{19.8} & \multicolumn{1}{c}{28.6} & 34.5          \\ 
\begin{tabular}[c]{@{}l@{}} $T \longrightarrow (ED, TD)$ \end{tabular}              & \multicolumn{1}{c}{\textbf{20.4}} & \multicolumn{1}{c}{\textbf{29.3}} & \textbf{34.9} \\ \hline
\end{tabular}
\caption{Ablation study results in GenKGC with different generating paths to distill corpus from LLMs. We conduct the experiment using FB15k-237N. We add the vallina GenKGC in the first row for comparison.}
\label{Tab: Ablation Study on Generating Path}
\end{table}

We investigate the efficacy of different context types in the distillation process by employing various generative paths.
As illustrated in Table~\ref{Tab: Ablation Study on Generating Path}, we initially explore the impact of entity description and triplet description when utilized separately as auxiliary corpora ($T \longrightarrow ED$ and $T \longrightarrow TD$).
The experimental findings underscore the critical roles played by both entity description and triplet description as distillation corpora, leading to noticeable enhancements in the performance of smaller KGC models.
Furthermore, we ascertain that our method's generating path $T \longrightarrow (ED, TD)$, which utilizes these two corpora, achieves more improvements by endowing the models with a more comprehensive and richer source of information.

To gain a comprehensive understanding of the effectiveness of our Contextualization Distillation, we also explored other alternative generative paths. While rationale distillation has demonstrated its potential in various NLP tasks~\cite{hsieh2023distilling,shridhar2023distilling}, our investigation delves into the $T \longrightarrow RA$ path, wherein we instruct the LLM to generate rationales for each training sample.
Although the model utilizing rationale distillation exhibits improved performance compared to the vanilla one, it falls short when compared with our Contextualization Distillation incorporating entity descriptions and triplet descriptions.
One plausible explanation for this disparity lies in the intrinsic nature of rationales, which tend to be intricate and structurally complex. This complexity can pose a greater challenge for smaller models to fully comprehend, in contrast to the more straightforward descriptive text utilized in our approach.

$T \longrightarrow ED \longrightarrow TD$ borrows the insight from Chain-of-CoT (CoT)~\cite{wei2022chain} that generates the content step by step.
Interestingly, our findings indicate that this multi-step generative path also yields suboptimal performance when compared to the single-step generative path. 
This discrepancy can be attributed to the text incoherence resulting from the concatenation of three segments of descriptions.
In light of the insights gained from these observations, we summarize our distillation guidance for KGC as follows: \textbf{smaller models can benefit more from comprehensive, descriptive and coherent content generated by LLMs}.

\subsection{Ablation Study on Descriptive Context}

\begin{table}[h]
\centering
\begin{tabular}{lccc}
\hline
\multirow{2}{*}{}                                               & \multicolumn{3}{c}{FN15k-237N}                                                                                              \\ \cline{2-4} 
                                                                                 & \multicolumn{1}{c}{H@1}           & \multicolumn{1}{c}{H@3}           & H@10          \\ \hline
GenKGC                                                                            & \multicolumn{1}{c}{18.7}          & \multicolumn{1}{c}{27.3}          & 33.7          \\
\hline
\begin{tabular}[c]{@{}l@{}}GenKGC\\ w/ Contextualization\\ w/ Wikipedia\end{tabular}                                                                           & \multicolumn{1}{c}{19.2}          & \multicolumn{1}{c}{27.9}          & 34.0          \\
\hline
GenKGC-CD                                                                            & \multicolumn{1}{c}{\textbf{20.4}}          & \multicolumn{1}{c}{\textbf{29.3}}          & \textbf{34.9}          \\
\hline
\end{tabular}
\caption{Ablation study results in GenKGC with descriptive context generated by our method and collected by~\citet{zhong2015aligning}.}
\label{Tab: Ablation Study on Descriptive Context}
\end{table}

In this section, we replace the auxiliary corpus used in the auxiliary task with the Wikipedia corpus collected by~\cite{zhong2015aligning} to study the effectiveness of the distillation.
As Table~\ref{Tab: Ablation Study on Descriptive Context} shows, while the auxiliary task with Wikipedia corpus improves the model’s performance, the overall enhancement is not as significant as that brought by our Contextualization Distillation. 
This further demonstrates \textbf{the corpus generated by large language models effectively tackles the limitations of the preceding corpus for KGC, resulting in more pronounced improvements for the KGC model}.

\subsection{Ablation Study on Generative KGC Models}
\label{Ablation Study on Generative KGC Models}

\begin{table}[]\small
\begin{tabular}{lcccc}
\hline
\multirow{2}{*}{}                                               & \multicolumn{4}{c}{FN15k-237N}                                                                                              \\ \cline{2-5} 
                                                                      & \multicolumn{1}{c}{MRR}           & \multicolumn{1}{c}{H@1}           & \multicolumn{1}{c}{H@3}           & H@10          \\ \hline
GenKGC                                                                & \multicolumn{1}{c}{-}             & \multicolumn{1}{c}{18.7}          & \multicolumn{1}{c}{27.3}          & 33.7          \\ 
\begin{tabular}[c]{@{}l@{}}w/ Reconstruction\end{tabular}    & \multicolumn{1}{c}{-}   & \multicolumn{1}{c}{19.4} & \multicolumn{1}{c}{28.2} & 34.2          \\ 
\begin{tabular}[c]{@{}l@{}}w/ Contextualization\end{tabular} & \multicolumn{1}{c}{-}             & \multicolumn{1}{c}{\textbf{20.4}} & \multicolumn{1}{c}{\textbf{29.3}} & \textbf{34.9} \\ \hline
KG-S2S                                                                & \multicolumn{1}{c}{35.4}          & \multicolumn{1}{c}{28.5}          & \multicolumn{1}{c}{38.8}          & 49.3          \\ 
\begin{tabular}[c]{@{}l@{}}w/ Reconstruction\end{tabular}    & \multicolumn{1}{c}{35.8}          & \multicolumn{1}{c}{\textbf{29.3}} & \multicolumn{1}{c}{38.9}          & 48.9          \\ 
\begin{tabular}[c]{@{}l@{}}w/ Contextualization\end{tabular} & \multicolumn{1}{c}{\textbf{35.9}} & \multicolumn{1}{c}{28.9}          & \multicolumn{1}{c}{\textbf{39.4}} & \textbf{50.2} \\ \hline
\end{tabular}
\caption{Ablation study results on GenKGC and KG-S2S with reconstruction and contextualization as the auxiliary task respectively. We conduct the experiment using FB15k-237N.}
\label{Tab: Ablation Study on Generative KGC Models}
\end{table}

In this section, we compare the effectiveness of reconstruction and contextualization in generative KGC models.
For GenKGC and KG-S2S, we employ the pre-trained tasks of their respective backbone models (BART for GenKGC and T5 for KG-S2S) as the reconstruction objective.
More details of our reconstruction implementation for generative KGC models can be found in Appendix~\ref{Implementation Details of Reconstruction for Generative KGC Models}.

Table~\ref{Tab: Ablation Study on Generative KGC Models} presents the ablation study results on FB15k-237N.
We find reconstruction is also effective in improving the performance of generative KGC models, showing that KGC models can consistently benefit from the descriptive context with different auxiliary tasks.
Comparing the two auxiliary tasks, models with contextualization outperform those with reconstruction on almost every metric, except for Hits@1 in KG-S2S.
This implies that \textbf{contextualization is a critical capability for generative KGC models to master for better KGC performance}.
Generative models have benefited more from the training of converting structural triplets into descriptive context than simply restoring the corrupted corpus.





\begin{table*}[h]
\centering

\begin{tabular}{lp{2.5in}p{2.5in}}
\hline
        & from Wikipedia~\cite{zhong2015aligning}                                                                                                                                                                                                             & Ours                                                                                                                                                                                                                                                                                                                                                                                                                                            \\ \hline
Head    & Ballard was a \textcolor[RGB]{76,175,80}{novelist}.                                                                                                                                                                                                      & J.G. Ballard (1930-2009) was an \textcolor[RGB]{76,175,80}{English writer}. He was born in Shanghai, China, and his early experiences there shaped his writing. His novels often \textcolor[RGB]{76,175,80}{explored themes of alienation, technology, and the future}...                                                               \\ \hline
Tail    & Shanghai is \textcolor[RGB]{244,67,54}{a 2010 American mystery/thriller neo-noir film} directed by Mikael Håfström, starring John Cusack and Gong Li... & Shanghai is \textcolor[RGB]{76,175,80}{a city in China}. It is one of the most populous cities in the world, and it is a major center of commerce and culture. Shanghai has a long history, and it has been \textcolor[RGB]{76,175,80}{home to many different cultures over the centuries}...                                                                                                                   \\ \hline
Triplet & In 1984, J.G. Ballard \textcolor[RGB]{244,67,54}{won broad, critical recognition} for the war novel Empire of the Sun, a \textcolor[RGB]{76,175,80}{semi-autobiographical} story of the experiences of a British boy during the Japanese \textcolor[RGB]{244,67,54}{occupation of} Shanghai.                         & Ballard \textcolor[RGB]{76,175,80}{was born in} Shanghai in 1930. He lived there until he was eight years old, when his family moved to England. Ballard's \textcolor[RGB]{76,175,80}{early experiences} in Shanghai had a profound impact on his writing... \\ \hline
\end{tabular}
\caption{Descriptive context of the triplet \emph{(J.G. Ballard, place\_of\_birth, Shanghai)}. The text in \textcolor[RGB]{76,175,80}{green} represents positive content and the text in \textcolor[RGB]{244,67,54}{red} represents negative content.}
\label{tab: Case Study}
\end{table*}

\subsection{Efficiency Analysis}

\begin{figure}[!t]
    \centering
    \includegraphics[width=7.7cm]{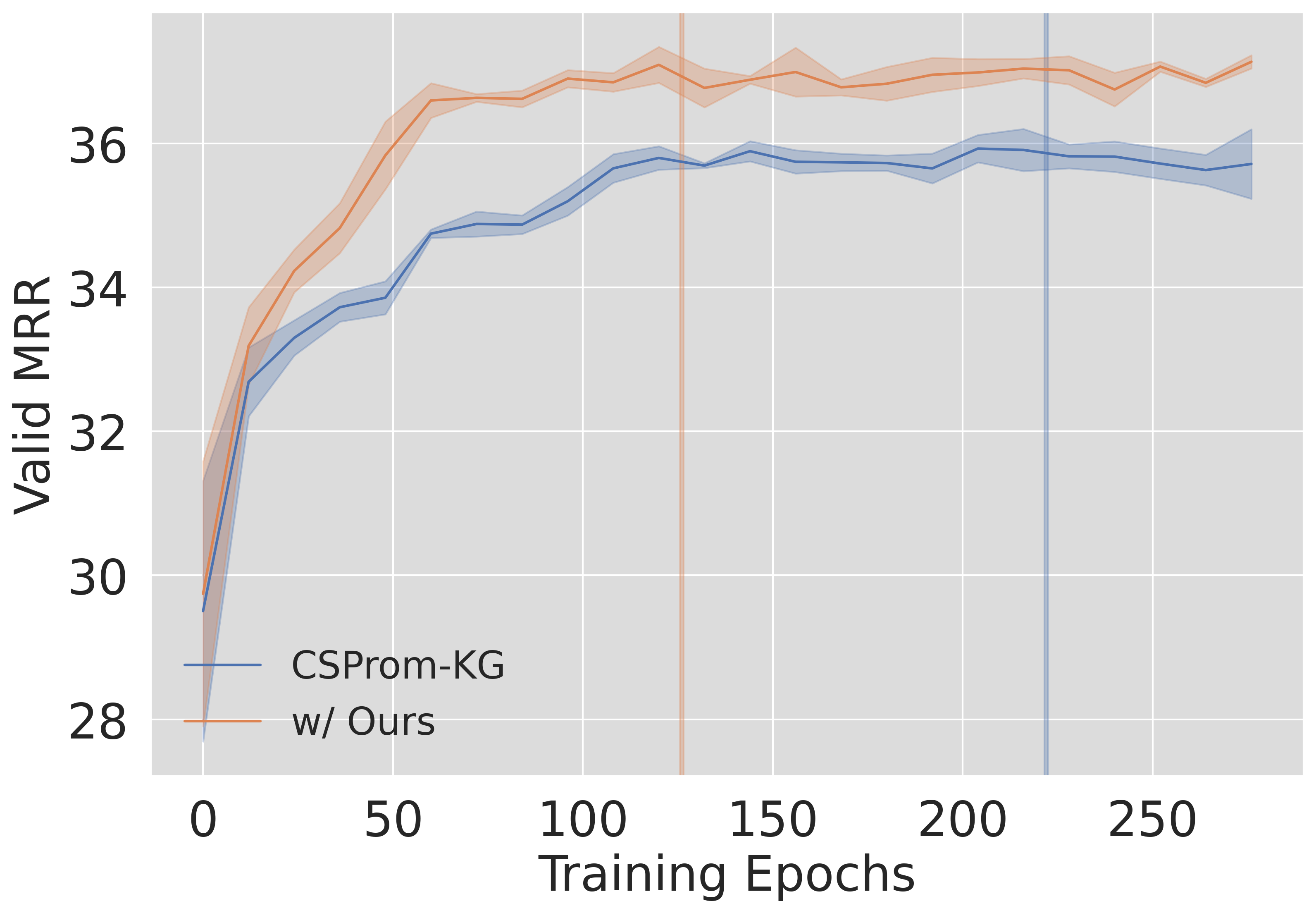}
    \caption{MRR scores on the validation set during the CSProm-KG training on FB15k-237N. We use thin bars to mark the epochs in which the models achieve the best performance in the validation set.}
    \label{efficiency}
\end{figure}

The additional training cost brought by the auxiliary distillation tasks may pose a potential constraint on our approach.
However, we also notice baseline models with our method coverage faster on the validation set.
Figure~\ref{efficiency} presents the validation MRR vs epoch numbers during the CSProm-KG training on FB15k-237N.
It is obvious that CSProm-KG with Contextualization Distillation achieves a faster convergence and attains the best checkpoint earlier (at around 125 epochs) compared to the variant without our method (at around 220 epochs).
This implies \textbf{auxiliary distillation loss can also expedite model learning in KGC}.
This trade-off between batch processing time and training steps ultimately results in a training efficiency comparable to that of the vanilla models.

\subsection{Case Study}

\begin{table}[]
\centering
\begin{tabular}{lp{1.8in}p{1.in}}
\hline
Query                  & \emph{(The Devil's Double, genre, ?)}                                                                                                                                \\ \hline
Ground Truth         & \emph{Biographical film}                                                                                                                                                   \\ \hline
Baseline    &\emph{ War film}                                                                                                                                                            \\ \hline
Ours         & \emph{Biographical film}                                                                                                                                                   \\ \hline
Our Context & The Devil's Double is a \textbf{biographical film} that tells the story of Latif Yahia, \textbf{a young Iraqi man who was forced to impersonate Saddam Hussein's son Uday Hussein}... \\ \hline
\end{tabular}
\caption{Case study on FB15K-237N with KG-S2S. we also let the model generate a descriptive context for each test sample. The text in \textbf{bold} represents informative content in the generated descriptive context.}
\label{tab:Case Study2}
\end{table}

We conduct a comparative analysis between the description corpus collected from Wikipedia~\cite{zhong2015aligning} and those generated using our method to show the advantage of our Contextualization Distillation more straightforwardly.
As presented in Table~\ref{tab: Case Study}, entity descriptions generated by the LLM effectively address the limitations issue and static shortcomings, resulting in more informative and accurate content.
Regarding the triplet description, although the ``semi-autobiographical'' used in ~\citet
{zhong2015aligning} somewhat implies J.G. Ballard's connection to Shanghai during his childhood, it still fails to express the semantics of ``\emph{place\_of\_birth}'' clearly.
In contrast, the descriptive context generated by our method provides a more elaborate and coherent contextualization of the ``\emph{place\_of\_birth}'' between ``\emph{J.G. Ballard}'' and ``\emph{Shanghai}''.
These comparisons highlight the effectiveness of our method in addressing the previous corpus' limitation.

Furthermore, We showcase how the auxiliary training with descriptive context enhances the baseline models.
Table~\ref{tab:Case Study2} presents the results of KG-S2S performance in a test sample of FB15k-237N, both with and without our contextualization distillation.
In this case, the vanilla KG-S2S wrongly predicts the genre of the film ``\emph{The Devil’s Double}'' as ``'\emph{War film}', whereas the KG-S2S trained with our auxiliary task correctly labels it as ``\emph{Biographical film}''.
Also, by making the model contextualize each triplet, we find the model with our method applied successfully captures many details about the movie, such as the genre and plot, and presents this information as fluent text.
In summary, \textbf{the model not only acquires valuable insights about the triplets but also gains the ability to adeptly contextualize this information through our Contextualization Distillation}.

Due to the space limitation, we put further analysis about LLMs' sizes in Appendix~\ref{Additional Analysis}.

\section{Conclusion}
In this work, we propose Contextualization Distillation, addressing the limitation of the existing KGC textual data by prompting LLMs to generate descriptive context.
To ensure the versatility of our approach across various PLM-based KGC models, we have designed a multi-task learning framework. Within this framework, we incorporate two auxiliary tasks, reconstruction and contextualization, which aid in training smaller KGC models in the informative descriptive context.
We conduct experiments on several mainstream KGC benchmarks and the results show that our Contextualization Distillation consistently enhances the baseline model's performance.
Furthermore, we conduct in-depth analyses to make the effect of our method more explainable, providing guidance on how to effectively leverage LLMs to improve KGC as well.
In the future, we plan to adapt our method to other knowledge-driven tasks, such as entity linking and knowledge graph question answering.

\section{Limitation}
Due to limitations in computing resources, we evaluate our method on two KGC datasets, while disregarding scenarios such as temporal knowledge graph completion~\cite{garcia2018learning}, few-shot knowledge graph completion~\cite{xiong2018one} and commonsense knowledge graph completion~\cite{li2022c3kg}. In future research, we plan to investigate the effectiveness of our method in border scenarios.

\bibliography{anthology,custom}

\appendix
\onecolumn

\section{Large Language Model Performance on KGC}

\label{Large Language Model Perform on KGC}

We follow~\citet{zhu2023llms} to assess the performance of directly instructing LLMs to perform KGC and Table~\ref{ICL sample} gives an example of our input to LLMs.
For PaLM, we utilize the API parameter ``candidate\_count'', while for ChatGPT, we use ``n'' to obtain multiple candidates, enabling the calculation of Hit@1, Hit@3, and Hit@10 metrics.
After obtaining the model's outputs, we use the Sentence-BERT~\cite{reimers2019sentence} to guarantee each output result matches a corresponding entity in the dataset's entity set.

Table~\ref{Additional LLMs performance on KGC} displays the additional experimental results for ChatGPT and PaLM2 across several KGC datasets.
Although LLMs demonstrate promising performance in a series of NLP tasks~\cite{liang2022holistic,yang2023new,chang2023survey} with various reasoning strategies~\cite{wei2022chain,wang2022self,li2023dail,tong2023eliminating}, they present a surprisingly poor performance in KGC with ICL.
It is evident that the performance of ICL of LLM falls short of KG-S2S's in every dataset.
One potential explanation for this subpar performance can be attributed to the phenomenon of hallucination in LLMs~\cite{ji2023survey,yang2023new}, leading to incorrect responses when the LLM encounters unfamiliar content.
Additionally,~\citet{li2023multi} exposes the ICL of LLMs' limitation in learning a domain-specific entity across the whole dataset, which provides another perspective to explain ICL's poor performance in KGC.

We also conducted an analysis of the influence of the number of demonstration samples.
As Table~\ref{demonstration number} shows, we find while the number of demonstrations increases, the performance of LLMs shows a corresponding improvement.
It appears that augmenting the number of demonstrations in the prompt could be a potential strategy for enhancing the capabilities of LLMs in KGC.
Nonetheless, it's essential to note that incorporating an excessive number of relevant samples as demonstrations faces practical challenges, primarily due to constraints related to input length and efficiency considerations.

\begin{table*}[h]
\centering
\begin{tabular}{lp{4.5in}p{2.9in}}
\hline

Triplet     & \emph{(Stan Collymore, play\_for, England national football team)}                                                                                                                                                                                                                                                                                                                                                                                                                                  \\ \hline
Tail Prompt & Predict the tail entity {[}MASK{]} from the given (Keko (footballer, born 1973), plays for, {[}MASK{]}) by completing the sentence "what is the plays for of Keko (footballer, born 1973)? The answer is ". The answer is UE Figueres, so the {[}MASK{]} is UE Figueres. Predict the tail entity {[}MASK{]} from the given (Stan Collymore, plays for, {[}MASK{]}) by completing the sentence "what is the plays for of Stan Collymore? The answer is ". The answer is                                \\ \hline
Head Prompt & Predict the head entity {[}MASK{]} from the given ({[}MASK{]}, plays for, UE Figueres) by completing the sentence "UE Figueres is the plays for of what? The answer is ". The answer is Keko (footballer, born 1973), so the {[}MASK{]} is Keko (footballer, born 1973). Predict the head entity {[}MASK{]} from the given ({[}MASK{]}, plays for, England national football team) by completing the sentence "England national football team is the plays for of what? The answer is ". The answer is \\ \hline
\end{tabular}
\caption{The prompt we use to directly leverage LLMs to perform KGC. Tail Prompt and Head Prompt mean the input to predict the missing tail and head entity respectively.}
\label{ICL sample}
\end{table*}

\begin{table*}[h]
\centering
\begin{tabular}{lccccccccc}
\hline
           & \multicolumn{3}{c}{ChatGPT}                                 & \multicolumn{3}{c}{PaLM2}                                                        & \multicolumn{3}{c}{KG-S2S}                                  \\ \cline{2-10} 
           & \multicolumn{1}{c}{H@1}  & \multicolumn{1}{c}{H@3}  & H@10 & \multicolumn{1}{c}{H@1}  & \multicolumn{1}{c}{H@3}  & H@8                       & \multicolumn{1}{c}{H@1}  & \multicolumn{1}{c}{H@3}  & H@10 \\ \hline
WN18RR     & \multicolumn{1}{l}{11.4} & \multicolumn{1}{l}{13.5} & 15.4 & \multicolumn{1}{l}{11.5} & \multicolumn{1}{l}{16.6} & \multicolumn{1}{l}{21.3} & \multicolumn{1}{c}{52.5} & \multicolumn{1}{c}{59.7} & 65.4 \\ 
FB15k-237  & \multicolumn{1}{c}{9.7}  & \multicolumn{1}{c}{11.2} & 12.4 & \multicolumn{1}{c}{11.5} & \multicolumn{1}{c}{16.6} & 21.7                      & \multicolumn{1}{c}{25.7} & \multicolumn{1}{c}{39.3} & 49.8 \\ 
FB15k-237N & \multicolumn{1}{c}{15.6} & \multicolumn{1}{c}{17.6} & 19.6 & \multicolumn{1}{c}{15.7} & \multicolumn{1}{c}{20.8} & 25.4                      & \multicolumn{1}{c}{28.5} & \multicolumn{1}{c}{38.8} & 49.3 \\ 
YAGO-3-10  & \multicolumn{1}{c}{4.5}  & \multicolumn{1}{c}{5.0}  & 5.4  & \multicolumn{1}{c}{6.4}  & \multicolumn{1}{c}{8.8}  & 11.4                      & \multicolumn{1}{c}{-}    & \multicolumn{1}{c}{-}    & -    \\ \hline
\end{tabular}
\caption{ChatGPT and PaLM2's results on other KGC datasets.}
\label{Additional LLMs performance on KGC}
\end{table*}

\begin{table*}[h]
\centering
\begin{tabular}{lccc}
\hline
             & \multicolumn{3}{c}{FB15k-237N}                                        \\ \cline{2-4}
             & \multicolumn{1}{c}{H@1}  & \multicolumn{1}{c}{H@3}  & \multicolumn{1}{c}{H@8} \\ \hline
PaLM2-1-shot & \multicolumn{1}{c}{15.7} & \multicolumn{1}{c}{20.8} & 25.4                     \\ 
PaLM2-2-shot & \multicolumn{1}{l}{16.9} & \multicolumn{1}{l}{22.1} & 26.8                     \\ 
PaLM2-4-shot & \multicolumn{1}{l}{17.7} & \multicolumn{1}{l}{23.1} & 27.9                     \\ \hline
\end{tabular}
\caption{Experiment results of the demonstration number's effect on LLMs when performing KGC.}
\label{demonstration number}
\end{table*}

\section{Details of Various KGC Pipelines}

\subsection{Discriminative KGC Pipelines}
\label{Discriminative KGC Pipelines}

KG-BERT~\cite{yao2019kg} is the first to propose utilizing PLMs for triplet modeling. It employs a special ``[CLS]'' token as the first token in input sequences. The head entity, relation, and tail entity are represented as separate sentences, with segments separated by [SEP] tokens. The input token representations are constructed by combining token, segment, and position embeddings. Tokens in the head and tail entity sentences share the same segment embedding, while the relation sentence has a different one. The input is fed into a BERT model, and the final hidden vector of the ``[CLS]'' token is used to compute triple scores. The scoring function for a triple (h, r, t) is calculated as $s = f(h, r, t) = sigmoid(CWT)$, where $s$ is a 2-dimensional real vector $s_{\tau0},s_{\tau1} \in [0,1]$ and $CWT$ is the embedding of the ``[CLS]'' token. Cross-entropy loss is computed using the triple labels and scores for positive and negative triple sets:
\begin{equation}
    \mathcal{L}_{kgc} = \sum_{\tau \in D^{+}+D^{-}} (y_{\tau}log(s_{\tau0})+(1-y_{\tau})log(s_{\tau1})),
\end{equation}
where $y_{\tau} \in \{0,1\}$ is the label of that triplet. The negative triplet $D^{-}$ is simply generated by replacing the head entity $h$ or tail entity $t$ in the original triplet $(h,r,t) \in D^{+}$.

CSProm-KG~\cite{chen2023dipping} combines PLM and traditional KGC models together to utilize both textual and structural information.
It first concatenates the entity description and relation description behind a sequence of conditional soft prompts as the input.
The input is then fed into a PLM, denoted as $P$, where the model parameters are held constant. Subsequently, CSProm-KG extracts embeddings from the soft prompts, which serve as the representations for entities and relations. These representations are then supplied as input to another graph-based KGC model, labeled as $G$, to perform the final predictions.
It also introduces a local adversarial regularization (LAR) method to enable the PLM $P$ to distinguish the true entities from $n$ textually similar entities $t^{l}$:
\begin{equation}
    \mathcal{L}_{l} = max(f(h,r,t), -\frac{1}{n}\sum_{i \in n}f(h,r,t_{i}^{l})+\gamma,0),
\end{equation}
where $\gamma$ is the margin hyper-parameter.
Finally, CSProm-KG utilizes the standard cross entropy loss with label smoothing and LAR to optimize the whole pipeline:
\begin{equation}
    \mathcal{L}_{c}= -(1-\phi)\cdot log\ p(t|h,r) - \frac{\phi}{|V|} \sum_{t^{'} \in V/t} log\ p(t^{'}h,r),
\end{equation}
\begin{equation}
    \mathcal{L}_{kgc} = \mathcal{L}_{c} + \beta\cdot\mathcal{L}_{l},
\end{equation}
where $\phi$ is the label smoothing value and $\beta$ is the LAR term weight.

\subsection{Generative KGC Pipelines}
\label{Generative KGC Pipelines}

In GenKGC~\cite{xie2022discrimination}, entities and relations are represented as sequences of tokens, rather than unique embeddings, to connect with pre-trained language models. For the triples $(e_i, r_j, e_k)$ with the tail entity $e_k$ missing, descriptions of $e_i$ and $r_j$ are concatenated to form the input sequence, which is then used to generate the output sequence. BART is employed for model training and inference, and a relation-guided demonstration approach is proposed for encoder training. This method leverages the fact that knowledge graphs often exhibit long-tailed distributions and constructs demonstration examples guided by the relation $r_j$. The final input sequence format is defined as: $x = [<BOS>,\ demonstration(r_j),\ <SEP>,\ d_{e_i},\ d{r_j},\ <SEP>]$, where $d_{e_i}$ and $d{r_j}$ are description of the head entity and relation respectively.
And $demonstration(r_j)$ means the demonstration examples with the relation $r_j$.
Given the input, the target of GenKGC in the decoding stage is to correctly generate the missing entity $y$, which can be formulated as:
\begin{equation}
    \mathcal{L}_{kgc} = -log\ p(e_K|x)
\end{equation}
Additionally, an entity-aware hierarchical decoding strategy has been proposed to improve the time efficiency.

Following them, KG-S2S~\cite{chen2022knowledge} adds the entity description in both the encoder and decoder ends, training the model to generate both the missing entity and its corresponding description.
It also maintains a soft prompt embedding for each relation to facilitate the model to distinguish the relations with similar surface meanings. 
Given the query $(e_i, r_j, e_k)$, the input $x$ and the label $y$ to predict the tail entity $e_k$ can be expressed as:
\begin{equation}
    x = [<BOS>, P_{e1},\ e_i,\ des_{e_i},\ P_{e1},\ <SEP>,\ P_{r1},\ r_j,\ P_{r2}],
\end{equation}
\begin{equation}
    y = [<BOS> e_k,\ des_{e_k}],
\end{equation}
where $des_{e}$ represents the entity description and $P$ here is the soft prompt embedding for entities or relations.
Additionally, it adopts a sequence-to-sequence dropout strategy by randomly masking some content in the entity description to avoid model overfitting in the training stage:
\begin{equation}
    x = RandomMask(x),
\end{equation}
and the total loss can be expressed as:
\begin{equation}
    \mathcal{L}_{kgc} = -log\ p(y|x)
\end{equation}

\section{Additional Implementation Details}
\label{Additional Implementation Details}

We show the detailed statistics of the KGC datasets we use in Table~\ref{Statistics}.
Table~\ref{hyper-parameter} displays the hyper-parameters we adopt for each baseline model and dataset.

\begin{table*}[h]
\centering
\begin{tabular}{cccccc}
\hline
Dataset    & \# Entity & \# Relation & \# Train  & \# Valid & \# Test  \\ \hline
WN18RR     & 40,943 & 11       & 86,835 & 3,034 & 3,134 \\ 
FB15k-237N & 14,541 & 93       & 87,282 & 7,041 & 8,226 \\ \hline
\end{tabular}
\caption{Statistics of the Datasets.}
\label{Statistics}
\end{table*}

\begin{table*}[h]
\centering
\begin{tabular}{llcccc}
\hline
model                      & dataset    & batch size & learning rate & epoch & $\alpha$ \\ \hline

\multirow{2}{*}{KG-BERT}   & WN18RR     & 32         & 5e-5          & 5     & 0.1   \\
                           & FB15k-237N & 32         & 5e-5          & 5     & 0.1   \\ \hline
\multirow{2}{*}{CSProm-KG} & WN18RR     & 128        & 5e-4          & 500   & 1.0   \\
                           & FB15k-237N & 128        & 5e-4          & 500   & 1.0   \\ \hline
\multirow{2}{*}{GenKGC}    & WN18RR     & 64         & 1e-4          & 10    & 1.0   \\
                           & FB15k-237N & 64         & 1e-4          & 10    & 1.0   \\ \hline
\multirow{2}{*}{KG-S2S}    & WN18RR     & 64         & 1e-3          & 100   & 0.5   \\
                           & FB15k-237N & 32         & 1e-3          & 50    & 0.5
\\ \hline
\end{tabular}
\caption{Details of hyper-parameter settings for each baseline and dataset.}
\label{hyper-parameter}
\end{table*}

\section{Implementation Details of Reconstruction for Generative KGC Models}

\label{Implementation Details of Reconstruction for Generative KGC Models}

In the case of GenKGC, we adhere to the denoising pre-training methodology used in BART~\cite{lewis2020bart}. This approach commences by implementing a range of text corruption techniques, such as token masking, sentence permutation, document rotation, token deletion, and text infilling, to shuffle the integrity of the initial text. The primary objective of BART's reconstruction task is to restore the original corpus from the corrupted text.

For KG-S2S, we follow the pre-training approach proposed by T5~\cite{raffel2020exploring}. This approach employs a BERT-style training objective and extends the concept of single token masking to encompass the replacement of text spans. In this process, we apply a 15\% corruption ratio for each segment, randomly substituting a span of text with a designated special token ``<extra\_id>''. Here we employ a span length of 3. The ultimate goal of T5's reconstruction task is to accurately predict the content associated with these special tokens.

\section{Analysis on LLMs' Sizes}
\label{Additional Analysis}

We conduct further analysis to validate the compatibility of our Contextualization Distillation with distillation models in various sizes.
We choose 3 smaller language models, GPT2, T5-base and T5-3B, each possessing comparable parameter counts to the KGC models we use (T5-base, BERT-base and BART-base). Additionally, we incorporated a larger language model, vicuna-7B, into our analysis. As the first step, we follow the method in Section 3.1 and instruct all these models to generate descriptive contexts for the triplet ”(J.G. Ballard| people, person, place\_of\_birth | Shanghai)”.

\begin{table*}[h]
\centering
\begin{tabular}{lp{4.5in}p{2.9in}}
\hline
Model      & Output                                                                                                                                          \\ \hline
GPT2-base  & relationship "people, person, place\_of\_birth". Please generate a paragraph to introduce "J.G.                                                 \\
T5-base    & The first paragraph should be a single sentence, with the following:\textbackslash{}n\textbackslash{}n"I am a person, person, place\_of\_birth. \\
T5-3B      & , person, place\_of\_birth | Shanghai) Contextualize: (J.G. Ballard|                                                                            \\
Vicuna-7B  & J.G. Ballard was a British novelist, short king, and essayist, best known for his dystopian and post-apocalyptic fiction...                     \\
PaLM2-540B & J.G. Ballard (1930-2009) was an English writer. He was born in Shanghai, China, and his...  \\ \hline                                             
\end{tabular}
\caption{Different models' contextualization output for the given triplet.}
\label{tab: model size}

\end{table*}

As shown in Table~\ref{tab: model size}, our observations reveal that the results produced by the three smaller language models (GPT-2, T5-base, and T5-3B) are subpar and irrelevant, indicating their incapacity to adhere to contextualization instructions effectively. By contrast, the context generated by Vicuna-7B is both fluent and informative, providing an accurate textual description of the entire triplet. So we conclude our first findings: \textbf{smaller language models, lacking the requisite capability to fully comprehend contextualization instructions and abstract triplets, are unsuitable as teacher models for our Contextualization Distillation}.

In the second step, we aim to investigate whether the context generated by smaller large language models would be beneficial for the KGC model. We follow exactly our method described in Section 3 and replace the PaLM2 with Vicuna-7B. We conducted an experiment in the FB15k-237N dataset with GenKGC as the KGC backbone model.

\begin{table*}[h]
\centering
\begin{tabular}{lccc}
\hline
\multirow{2}{*}{}                                               & \multicolumn{3}{c}{FN15k-237N}                                                                                              \\ \cline{2-4} 
                                                                                 & \multicolumn{1}{c}{H@1}           & \multicolumn{1}{c}{H@3}           & H@10          \\ \hline
GenKGC                                                                            & \multicolumn{1}{c}{18.7}          & \multicolumn{1}{c}{27.3}          & 33.7          \\
\hline
\begin{tabular}[c]{@{}l@{}}GenKGC-CD\\ w/ Vicuna-7B\end{tabular}                                                                           & \multicolumn{1}{c}{19.9}          & \multicolumn{1}{c}{28.6}          & 34.6          \\
\hline
\begin{tabular}[c]{@{}l@{}}GenKGC-CD\\ w/ PaLM2-540B\end{tabular}                                                                            & \multicolumn{1}{c}{\textbf{20.4}}          & \multicolumn{1}{c}{\textbf{29.3}}          & \textbf{34.9}          \\
\hline
\end{tabular}
\caption{Comparison between our method using Vicuna-7B and PaLM2-540B.}
\label{Tab: vicuna 7B}
\end{table*}

As depicted in Table~\ref{Tab: vicuna 7B}, our Contextualization Distillation with Vicuna-7B remains effective in enhancing the KGC model, albeit not to the extent observed with CD utilizing PaLM2. This leads us to the conclusion that \textbf{Contextualization Distillation is also compatible with large language models with fewer parameters, even as small as 7B in size}. In the future, we will continue to explore the impact of different language model sizes (such as 13B and 30B) on our method.

\section{Additional Case Study}
\label{Additional Case Study}

In this section, we provide detailed examples to illustrate the input and output of each generating path we adopt in the descriptive context/ rationale extraction stage.
We present examples in Table~\ref{generating path: ours},~\ref{generating path: description},~\ref{generating path: relation},~\ref{generating path: step},~\ref{generating path: rationale}.

\begin{table*}[h]
\centering
\begin{tabular}{lp{4.5in}p{2.9in}}
\hline
Triplet & (J.G. Ballard, place\_of\_birth, Shanghai)                                                                                                                                                                                                                                                                                                                                                                                                                                                                                                                                                                                                                                                                                                                                                                                                                                                                                                                                                                                                                                                                                                                 \\ \hline
Prompt  & Given a triplet (J.G. Ballard| people, person, place\_of\_birth | Shanghai),please generate a paragraph to introduce "J.G. Ballard" and "Shanghai" and reflect their relationship "people, person, place\_of\_birth".                                                                                                                                                                                                                                                                                                                                                                                                                                                                                                                                                                                                                                                                                                                                                                                                                                                                                                                                                                  \\ \hline
Output  & J.G. Ballard (1930-2009) was an English writer. He was born in Shanghai, China, and his early experiences there shaped his writing. His novels often explored themes of alienation, technology, and the future. Ballard's work has been praised for its originality and its insights into modern society. He is considered one of the most important writers of the 20th century. Shanghai is a city in China. It is one of the most populous cities in the world, and it is a major center of commerce and culture. Shanghai has a long history, and it has been home to many different cultures over the centuries. The city is known for its skyscrapers, its nightlife, and its vibrant atmosphere. J.G. Ballard was born in Shanghai in 1930. He lived there until he was eight years old, when his family moved to England. Ballard's early experiences in Shanghai had a profound impact on his writing. His novels often explore themes of alienation, technology, and the future. Ballard's work has been praised for its originality and its insights into modern society. He is considered one of the most important writers of the 20th century. \\ \hline
\end{tabular}
\caption{Descriptive context obtained from the generating path $T \longrightarrow (ED, TD)$.}
\label{generating path: ours}
\end{table*}

\begin{table}[h]
\centering
\begin{tabular}{lp{4.5in}p{2.9in}}
\hline
Triplet & (J.G. Ballard, place\_of\_birth, Shanghai)                                                                                                                                                                                                                                                                                                                                                                                                                \\ \hline
Prompt  & Given a triplet (J.G. Ballard | people, person, place\_of\_birth | Shanghai), please generate a short paragraph to introduce "J.G. Ballard" and "people, person, place\_of\_birth". "J.G. Ballard" and "people, person, place\_of\_birth" must be concluded in the generated text.                                                                                                                                                                                                                                                                                                  \\ \hline
Output  & J.G. Ballard (1930-2009) was a British novelist, short story writer, and essayist. Born in Shanghai, China, he was raised there until the age of eight, when his family returned to England. Ballard's early experiences in China had a profound impact on his writing, and many of his novels and stories explore themes of alienation, isolation, and dystopia. He is considered one of the most important science fiction writers of the 20th century. \\ \hline
\end{tabular}
\caption{Descriptive context obtained from the generating path $T \longrightarrow ED$.}
\label{generating path: description}
\end{table}

\begin{table}[h]
\centering
\begin{tabular}{lp{4.5in}p{2.9in}}
\hline
Triplet & (J.G. Ballard, place\_of\_birth, Shanghai)                                                                                                                                                                                                                                                                                                                                                    \\ \hline
Prompt  & Given a triplet (J.G. Ballard | people, person, place\_of\_birth | J.G. Ballard), please generate a paragraph to reflect their relationship people, person, place\_of\_birth. "J.G. Ballard" and "Shanghai" must concluded in the generated text.                                                                                                                                                                                                                                       \\ \hline
Output  & J.G. Ballard was born in Shanghai, China on November 15, 1930. He was a British writer, best known for his science fiction novels and short stories. His work often explored themes of alienation, technology, and the human condition. Ballard's books have been translated into more than 30 languages and have sold over 15 million copies worldwide. He died in London on April 19, 2009. \\ \hline
\end{tabular}
\caption{Descriptive context obtained from the generating path $T \longrightarrow TD$.}
\label{generating path: relation}
\end{table}

\begin{table}[h]
\centering
\begin{tabular}{lp{4.5in}p{2.9in}}
\hline
Triplet        & (J.G. Ballard, place\_of\_birth, Shanghai)                                                                                                                                                                                                                                                                                                                                                         \\ \hline
Prompt-Head    & Given a triplet (J.G. Ballard | people, person, place\_of\_birth | Shanghai), please generate a short paragraph to introduce "J.G. Ballard"                                                                                                                                                                                                                                                                                                   \\ \hline
Prompt-Tail    & Given a triplet (J.G. Ballard | people, person, place\_of\_birth | Shanghai), please generate a short paragraph to introduce "Shanghai"                                                                                                                                                                                                                                                                                                   \\ \hline
Prompt-Triplet & Given a triplet (J.G. Ballard | people, person, place\_of\_birth | Shanghai) and the description of the two entities J.G. Ballard: \textless{}Output-Head\textgreater Shanghai: \textless{}Output-Tail\textgreater please generate a short paragraph to reflect their relationship people, person, place\_of\_birth. "J.G. Ballard" and "Shanghai" must be concluded in the generated text.                                                                                                   \\ \hline
Output-Head    & J. G. Ballard was born in Shanghai and is a British writer. He is best known for his novels about science fiction and post-apocalyptic fiction. Some of his most famous works include "The Drowned World", "The Crystal World", and "Empire of the Sun".                                                                                                                                           \\ \hline
Output-Tail    & Shanghai is a metropolis on the east coast of China. It is the largest city in China and one of the most populous cities in the world. Shanghai is a major economic and cultural center, and is home to many skyscrapers and other landmarks. The city is also known for its nightlife and its vibrant street life. J.G. Ballard, the author of "Empire of the Sun", was born in Shanghai in 1930. \\ \hline
Output-Triplet & J. G. Ballard was born in Shanghai, China in 1930. He is a British writer best known for his novels about science fiction and post-apocalyptic fiction. Some of his most famous works include "The Drowned World", "The Crystal World", and "Empire of the Sun".                                                                                                                                   \\ \hline
\end{tabular}
\caption{Descriptive context obtained from the generating path $T \longrightarrow ED \longrightarrow TD$. <Output-Tail> and <Output-head> refer to the tail description and head description generated by the LLM in previous steps.}
\label{generating path: step}
\end{table}

\begin{table}[h]
\centering
\begin{tabular}{lp{4.5in}p{2.9in}}
\hline
Triplet     & (J.G. Ballard, place\_of\_birth, Shanghai)                                                                                                                                                                                                                                                                                                                                                                                                                                                                                                                                                                                                                                                       \\ \hline
Prompt-Tail & Given a triplet (J.G. Ballard | people, person, place\_of\_birth | {[}MASK{]}), please generate rationale to explain why {[}MASK{]} should be Shanghai                                                                                                                                                                                                                                                                                                                                                                                                                                                                                                                                                                   \\ \hline
Prompt-Head & Given a triplet ({[}MASK{]} | people, person, place\_of\_birth | J.G. Ballard), please generate rationale at least 100 words to explain why {[}MASK{]} should be J.G. Ballard                                                                                                                                                                                                                                                                                                                                                                                                                                                                                                                                                                   \\ \hline
Output-Tail & The rationale for {[}MASK{]} should be Shanghai is that:1. J.G. Ballard was born in Shanghai, China.2. Ballard's father was a British engineer who worked in China.3. Ballard spent his early childhood in Shanghai.4. Ballard's experiences in Shanghai influenced his writing.5. Ballard's novels and short stories often feature themes of alienation, isolation, and dystopia.6. Ballard's work has been praised for its originality, its insights into human nature, and its depiction of the future.In conclusion, the rationale for {[}MASK{]} should be Shanghai is that Ballard was born and raised there, and his experiences in Shanghai shaped his writing.                          \\ \hline
Output-Head & J.G. Ballard was born in Shanghai on 15 November 1930. His father was a British businessman and his mother was a Chinese woman. Ballard lived in Shanghai until the Japanese invasion in 1937, when he and his family were evacuated to England. Ballard's experiences in Shanghai during the war had a profound impact on his writing, and many of his novels are set in dystopian cities that are reminiscent of Shanghai. Ballard's work is often characterized by its bleak vision of the future, but it also contains a strong element of hope. Ballard believed that the future could be a place of beauty and wonder, but only if we are willing to confront the challenges that we face. \\ \hline
\end{tabular}
\caption{Rationale obtained from the generating path $T \longrightarrow RA$}
\label{generating path: rationale}
\end{table}

\end{document}